\documentclass{llncs}
\usepackage{url}
\usepackage{amsmath,amssymb,amsfonts}
\usepackage{algorithmic}
\usepackage{graphicx,multirow}
\usepackage{textcomp}
\usepackage{float}
\usepackage{xcolor}
\graphicspath{{figs/}}

\begin{document}

\title{Explaining Neural Networks by Decoding Layer Activations}
\author{Johannes Schneider\inst{1} \and Michalis Vlachos \inst{2}}
\institute{Institute of Information Systems, University of Liechtenstein, Liechtenstein  \and Department of Information Systems, HEC, University of Lausanne, Switzerland  }
% \author{Johannes Schneider,\\
% University of Liechtenstein,\\
% Principality of Liechtenstein\\
% johannes.schneider@uni.li\\
% \And
% Michalis Vlachos,\\
% University of Lausanne,\\
% Switzerland}

\maketitle

\begin{abstract} %, such as the training data
We present a `CLAssifier-DECoder' architecture (\emph{ClaDec}) which facilitates the comprehension of the output of an arbitrary layer in a neural network (NN). It uses a decoder to transform the non-interpretable representation of the given layer to a representation that is more similar to the domain a human is familiar with.  In an image recognition problem, one can recognize what information is represented by a layer by contrasting reconstructed images of \emph{ClaDec} with those of a conventional auto-encoder(AE) serving as reference. We also extend \emph{ClaDec} to allow the trade-off between human interpretability and fidelity. We evaluate our approach for image classification using Convolutional NNs. We show that reconstructed visualizations using encodings from a classifier capture more relevant information for classification than conventional AEs. Relevant code is available at \url{https://github.com/JohnTailor/ClaDec} .
%In alignment with our theoretical motivation, %
%The qualitative evaluation highlights that reconstructed images (of the network to be explained) tend to replace specific objects with more generic object templates and provide smoother reconstructions.
\end{abstract}
\section{Introduction} %,schneider2019pers
%Tacit or implicit knowledge, refers to the knowledge that can  \textit{explain} how to perform a task. Such knowledge is difficult to transfer not only among humans but also between humans and NNs. Similarly, Explaining predictive models is important for many reasons, including: a) debugging or improving models, b) fulfilling legal obligations such as the "right to explanation" as crystallized in the European GDPR data privacy law, c) increasing trust in models. Thus, it is not surprising that explaining NNs has received a lot of attention \cite{adadi2018peeking}. 
Understanding a NN is a multi-faceted problem, ranging from understanding single decisions, single neurons and single layers, up to explaining  complete models. In this work, we are interested in better understanding the  decision of a NN with respect to one or several user-defined layers that originate from a complex feature hierarchy, as commonly found in deep learning models. In a layered model, each layer corresponds to a transformed representation of the original input. Thus, the NN succinctly transforms the input into representations that are more useful for the task at hand, such as classification. From this point of view, we seek to answer the question: \textit{``Given an input $X$, what does the representation $L(X)$ produced in a layer $L$ tell us about the decision and about the network?''}. To address this question, we propose a classifier-decoder architecture called \emph{ClaDec}. It uses a decoder to transform the representation $L(X)$ produced by a layer $L$ of the classifier, with the goal to explain that layer via a human understandable representation, i.e., one that is similar to the input domain. The layer in question provides the ``code''  that is fed into a decoder. The motivation for this architecture stems from the observation that AE architectures are good at (re)constructing high-dimensional data from a low-dimensional representation.  The idea behind this, stems from the observation that the classifier to be explained is expected to encode faithfully aspects relevant to the classification and ignore input information that does not impact decisions. Therefore, use of a decoder can lead to accurate reconstruction of parts and attributes of the input that are essential for classification. In contrast, inputs that have little or no influence to the classification will be reconstructed at lower fidelity. Attributes of an input might refer to basic properties such as color, shape, sharpness but also more abstract, higher-level concepts. That is, reconstructions of higher-level constructs might be altered to be more similar to prototypical instances. %\textbf{We provide a theoretically founded motivation in the Appendix.}
 %as shown in the methods overview in Figure \ref{fig:met}. We shall discuss details to the various approaches in the related work. Furthermore, to compute explanations we only utilize the latent code and no other information such as gradients. Multiple explanation methods fall short in several ways such as reacting to irrelevant parts of the input \cite{adebayo2018sanity} and being sensitive to factors not contributing to model predictions \cite{kin19}. 

%, potentially even depending on the individual receiving an explanation \cite{schneider2019pers}

Explanations should fulfill many partially conflicting objectives. We are interested in the trade-off between fidelity (How accurately does the explanation express the model behavior?) and interpretability (How easy is it to make sense of the explanation?). While these properties of explanations are well-known, existing methods typically do not accommodate adjusting this trade-off. In contrast, we propose an extension of our base architecture \emph{ClaDec} by adding a classification loss. It allows to balance between producing reconstructions that are similar to the inputs, i.e., training data that a user is probably more familiar with (easier interpretation), and reconstructions that are strongly influenced by the model to explain (higher fidelity) but may deviate more from what the user knows or has seen.
\begin{figure}
  \centering
  \vspace{-18pt}
  \includegraphics[width=0.8\linewidth]{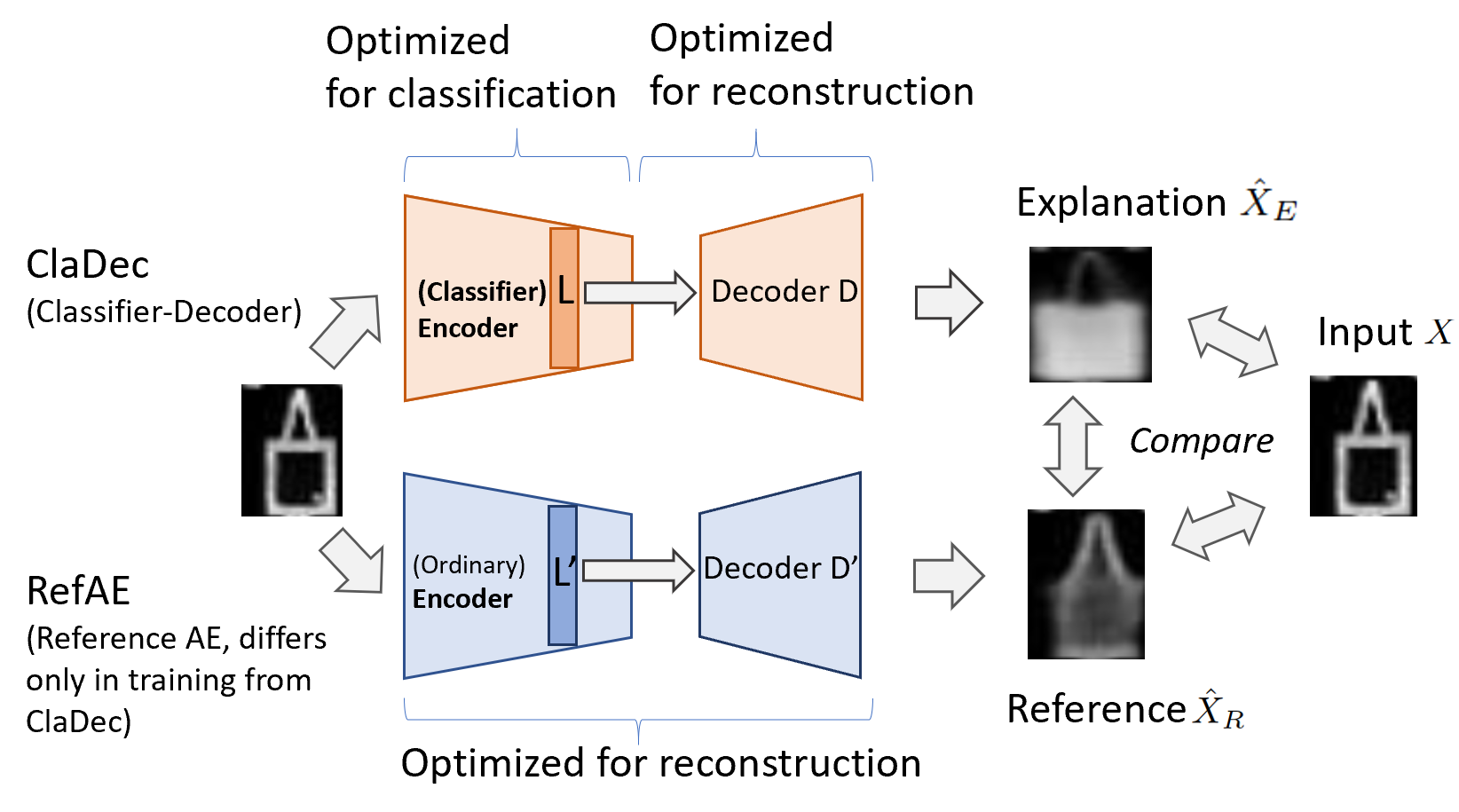}
  \vspace{-9pt}
  \caption{Basic architecture of \emph{ClaDec} and \emph{RefAE} and explanation process} \label{fig:arch}
  \vspace{-18pt}
\end{figure}
Our approach relies on an auxiliary model, a decoder, to provide explanations. Similar to other methods that use auxiliary or proxy models, e.g., to synthesize inputs \cite{ngu16} or approximate model behavior \cite{ribeiro2016should}, we face the problem that explanation fidelity may be negatively impacted by a poor auxiliary model. That is, reconstructions produced by AEs (or GANs) might suffer from artifacts. For example, AEs are known to produce images that might appear more blurry than real images. People have noticed that GANs can produce clearer images but they may suffer from other artifacts as shown in \cite{ngu16}.  Neglecting that the explainability method might introduce artifacts can have an adverse impact on understandability and even lead to wrong conclusions on model behavior. When looking at the reconstruction, a person not familiar with such artifacts might not attribute the distortion to the auxiliary model being used but she might believe that it is due to the model to be explained. While evaluation of explainability methods has many known open questions \cite{yang19}, this is the first work that has made this observation. 

To avoid any wrongful perceptions with respect to artifacts in reconstruction, we suggest to compare outcomes of auxiliary models to a reference architecture. We employ an auto-encoder \emph{RefAE} with the exact same architecture as \emph{ClaDec} to generate outputs for comparison as shown in Figure \ref{fig:arch}. The encoder of  \emph{RefAE} is not trained for classification, but the \emph{RefAE} model optimizes the reconstruction loss of the original inputs as any conventional AE. Therefore, only the differences visible in the reconstructions of  \emph{RefAE} and \emph{ClaDec} can be attributed to the model to be explained. 
%\\ %despite containing less information on the inputs.\\
The proposed comparison to a reference model can also be perceived as a rudimentary sanity check, i.e., if there are no differences then either the explainability method is of little value or the objective of the model to be explained is similar to that of the reference AE, as we shall elaborate more in our theoretical motivation. We believe that such sanity checks are urgently needed, since multiple explanation methods have been scrutinized for failing ``sanity'' checks and simple robustness properties \cite{adebayo2018sanity,kin19,gho19}. For that reason, we also introduce a sanity check that formalizes the idea that inputs plus explanations should lead to better performance on downstream tasks than inputs alone. In our context, we even show that auxiliary classifiers trained on either reconstructions from \emph{RefAE} or \emph{ClaDec} perform better on the latter, although the reference AE leads to reconstructions that are closer to the original inputs. Thus, the reconstructions of \emph{ClaDec} are more amendable for the task to be solved. Overall, we make the following \textbf{contributions}:\\
%\begin{itemize} %, theoretically grounded
    %\item 
\noindent i) We present a novel method to understand layers of NNs. It uses a decoder to translate non-interpretable layer outputs into a human understandable representation. It allows to trade interpretability and fidelity.\\ % It leads to easy to interpret explanations that also allow to derive general statements upon model behavior that are more difficult to derive using other methods such as saliency maps based on perturbations or gradients.
\noindent ii) We introduce a method dealing with artifacts created by auxiliary models (or proxies) through comparisons with adequate references. This includes evaluation of methods.
%\noindent iii) We contribute to the evaluation of explainability methods based on auxiliary models. by formalizing the evaluation of different objectives of explanations, such as fidelity and interpretability. 
    % focusing on image classification
    %\item Performing qualitative evaluation and quantitative evaluation including sanity checks for the proposed method.
%\end{itemize}

%In this work, we address the question ``How does a classifier represent an input at a specific layer?'' The question is easy to answer, if we aim for fidelity: It's just the activations of the neurons of that layer. However, this information is not interpretable by a human. Therefore, we propose to map it back to a space that is meaningful to the human, ie. using a decoder.

%Explanations are often difficult to understand and to make sense of. One reason might be that NNs are complex and exhibit unintiutive traits. For example, a single pixel can be changed to change among class (adversarial attacks). 

%That is, we aim at learning p(X|f(X)) 

\section{Method and Architecture} %Figure \ref{fig:arch} outlines our base architecture. % that might also be visible for an encoder that is optimized only towards reconstruction

The \emph{ClaDec} architecture is shown on the top portion of Figure \ref{fig:arch}. It consists of an encoder and a decoder reconstructing the input. The encoder is made of all layers of a classifier up to a user-specified layer $L$. The entire classifier has been trained beforehand to optimize classification loss. Its parameters remain unchanged during the explanation process. To explain layer $L$ of the classifier for an input $X$, we use the activations of layer $L(X)$. The activations $L(X)$ are provided to the decoder. The decoder is trained to optimize the reconstruction loss with respect to the original inputs $X$. The \emph{RefAE} architecture is identical to \emph{ClaDec}. It differs only in the training process and the objective. For the reference AE, the encoder and decoder are trained jointly to optimize the reconstruction loss of inputs $X$. In contrast, the encoder is treated as fixed in \emph{ClaDec}. Once the training of all components is completed, explanations can be generated without further need for optimization. That is, for an input $X$, \emph{ClaDec} computes the reconstruction $\hat{X}_E$ serving together with the original input and the reconstruction from \emph{RefAE} as the explanation. 

\begin{figure}
  \centering
  \vspace{-9pt}
  \includegraphics[width=0.8\linewidth]{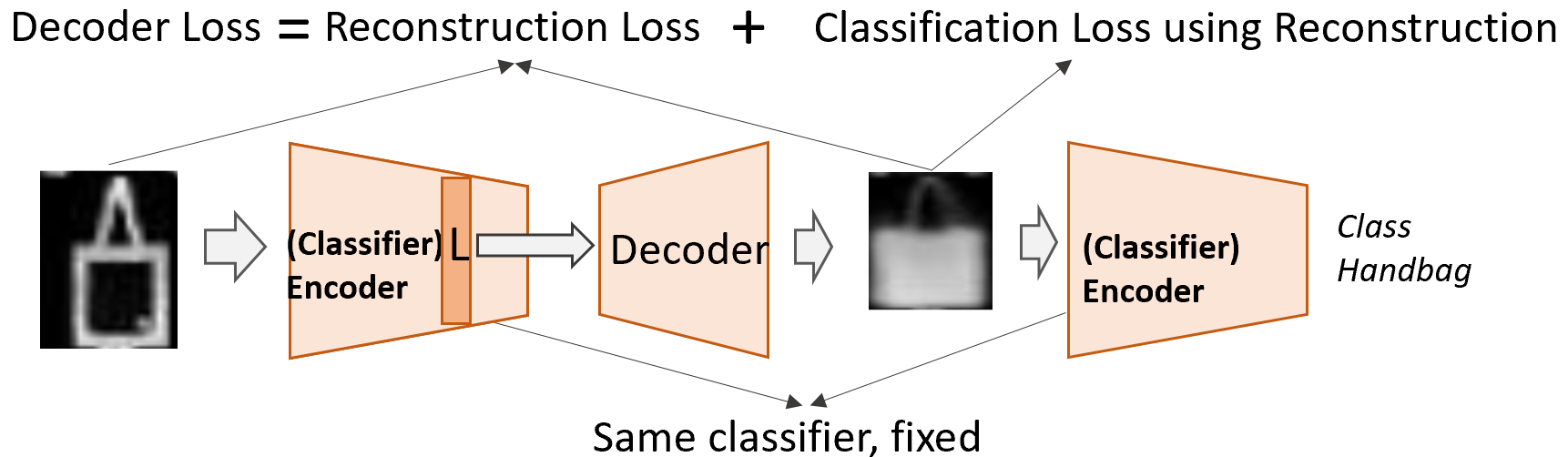}
  \vspace{-5pt}
  \caption{Extension of the \emph{ClaDec} architecture. The decoder is optimized for reconstruction and classification loss} \label{fig:deco}
  \vspace{-12pt}
\end{figure}

However, comparing the reconstruction $\hat{X}_E$ to the input $X$ may be difficult and even misleading, since the decoder can introduce distortions. Image reconstruction in general by AEs or GANs is not perfect. Therefore, it is unclear, whether the differences between the input and the reconstruction originate from the encoding of the classifier or the inherent limitations of the decoder. This problem exists in other methods as well, e.g. \cite{ngu16}, but it has been ignored. Thus, we propose to use both the \emph{RefAE} (capturing unavoidable limitations of the model or data) and \emph{ClaDec} (capturing model behavior). The evaluation proceeds by comparing the reconstructed ``reference'' from \emph{RefAE}, the explanation from \emph{ClaDec} and the input. Only differences between the input and the reconstruction of \emph{ClaDec} that do not occur in the reconstruction of the reference can be attributed to the classifier.
%Figure \ref{fig:deco} provides more details and an extension allowing to specify how close reconstructions should be to inputs minimizing classification loss. % It also highlights in more detail the setup, when the goal is to explain the output of a specific layer $L$. That is the activation of this layer is used to reconstruct the explanation $\hat{X}_E$  using the decoder. The reconstructed image $\hat{X}_E$ is then used to compute the classification loss.  The extension in Figure \ref{fig:deco}
Figure \ref{fig:deco} shows an extension of the base architecture of \emph{ClaDec} (Figure \ref{fig:arch}) using a second loss term for the decoder training. It is motivated by the fact that \emph{ClaDec} seems to yield reconstructions that capture more aspects of the input domain than of the classifier. That is, reconstructions might be easy to interpret, but in some cases it might be preferable to allow for explanations that are more fidel, ie. capturing more aspects of the model that should be explained.
%In the evaluation, we shall also consider a variant that replaces the classification loss with a reconstruction loss of the layer $L$, ie. the difference between $L$ and $\hat{L}$, where $L$ are the activations based on the original input and $\hat{L}$ of the reconstructed input.

More formally, for an input $X$, a classifier $C$ (to be explained) and a layer $L$ to explain, let $L(X)$ be the activations of layer $L$ for input $X$, and $Loss(CL(X),Y)$ the classification loss of $X$ depending on the true classes $Y$. The decoder $D$ transforms the representation $L(X)$ into the reconstruction $\hat{X}$.
For \emph{ClaDec} the decoder loss is:
\begin{equation} 
%\vspace{-3pt}
\footnotesize
\begin{aligned}	%_{ClaDec}
Loss(X)&:= (1-\alpha)\cdot \sum_i (X_i-\hat{X}_{E,i})^2 + \alpha\cdot Loss(CL(\hat{X}_{E}),Y)\\
&\text{ with } \hat{X}_{E}:=D(L(X)) \text{ and }\alpha \in[0,1] 
 \end{aligned}	\label{eq:alpha}
%\vspace{-3pt}
\end{equation}
The trade-off parameter $\alpha$ allows to control whether reconstructions $\hat{X}_{E}$ are more similar to inputs with which the domain expert is more familiar, or  reconstructions that are more shaped by the classifier and, thus, they might look more different than training data a domain expert is familiar with. For reconstructions $X_{R,i}$ of \emph{RefAE} the loss is only the reconstruction loss $\sum_i (X_i-\hat{X}_{R,i})^2$.
%For the reference AE \emph{RefAE}, the loss of the decoder $D'$ is simpler, ie. it is merely the reconstruction loss:
%\begin{small}
%$$Loss_{RefAE}(X):= \sum_i (X_i-\hat{X}_{R,i})^2 \text{ with } \hat{X}_R:=D'(L(X))$$
%\end {small}
\section{Theoretical Motivation of \emph{ClaDec}} %of the Classifier-Decoder  %https://math.stackexchange.com/questions/245141/do-the-real-numbers-and-the-complex-numbers-have-the-same-cardinality
\begin{figure}
  \centering
  \vspace{-3pt}
  \includegraphics[width=0.6\linewidth]{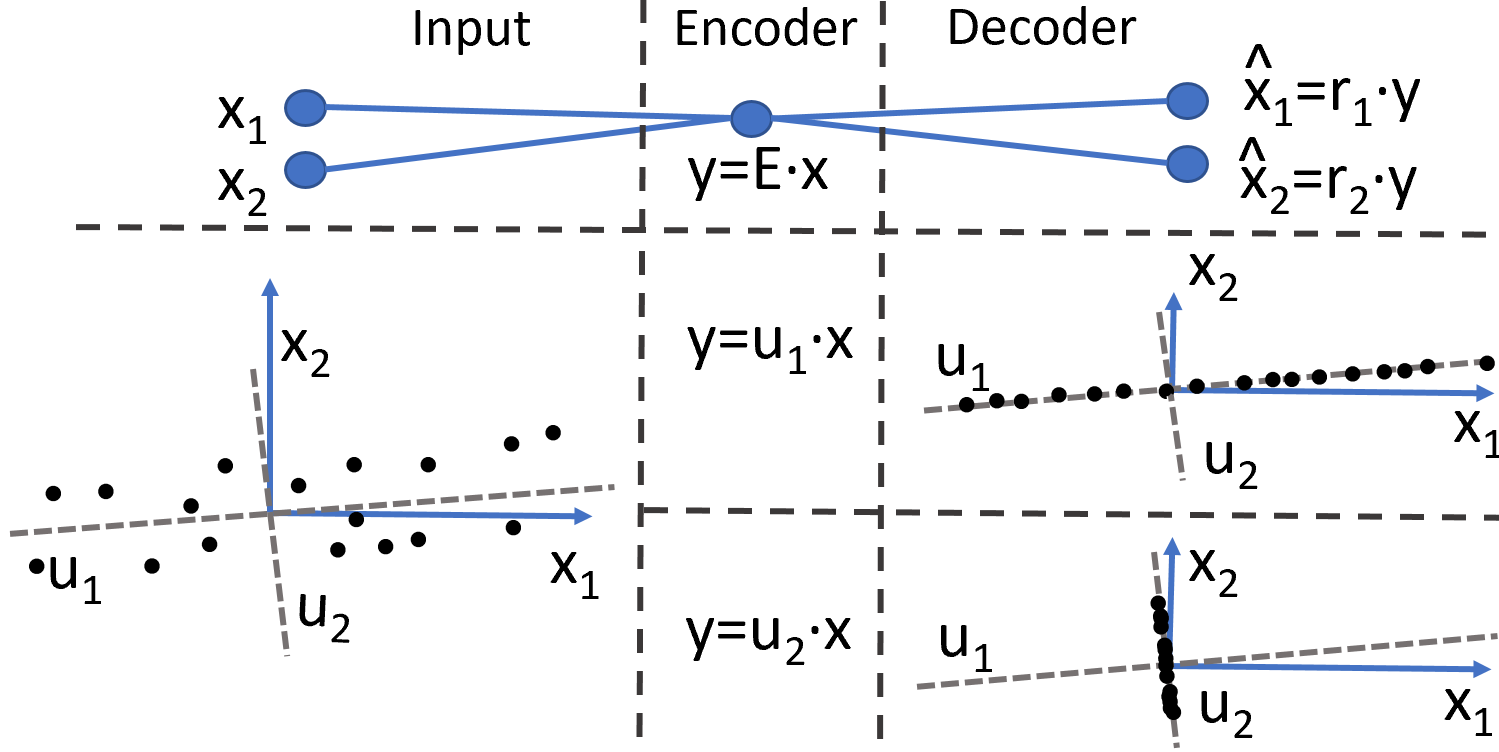}
  \vspace{-3pt}
  \caption{An AE with optimal encoder $y=u_1\cdot x$ (and decoder) captures more information than any other encoder. But a regression/classification model serving as encoder, eg. $y=u_2\cdot x$, combined with an optimized decoder, might capture some input attributes more accurately, eg. $x_2$.} \label{fig:the}
  \vspace{-3pt}
\end{figure}

We provide rational for reconstructing explanations using a decoder from a layer of a classifier that should be explained, and comparing it to the output of a conventional AE, ie. \emph{RefAE} (see Figure \ref{fig:arch}). AEs perform a transformation of inputs to a latent space and then back to the original space. This comes with information loss on the original inputs because reconstructions are typically not identical to inputs.  %It may appear that this information loss is due to forcing high-dimensional data to be represented in a low dimensional space. However, as claimed in \cite{goo16}(p.505), a non-linear encoder and decoder (theoretically) only require a single dimension to encode arbitrary information without any loss. The deeper mathematical reason is that a dimension $d$ is a real number, ie. $d \in \mathbb{R}$ and real numbers are uncountable infinite. Thus, there are (more than) enough options to encode an infinite amount of inputs.% The latter implies that using (finitely) more dimensions cannot increase the cardinality of the space, ie. $|\mathbb{R}|^k=|\mathbb{R}|$ for constant $k$.\footnote{Short proof: $|\mathbb{R}|^k=(2^{\aleph_0})^k=2^{k\aleph_0}=2^{\aleph_0}=|\mathbb{R}|$, where $\aleph_0$ is the aleph null symbol denoting cardinality of the set of all natural numbers.}
To provide intuition, we focus on a simple architecture with a linear encoder (consisting of a linear model that should be explained), a single hidden unit and a linear decoder as depicted in Figure \ref{fig:the}.
An AE, ie. the reference AE  \emph{RefAE}, aims to find an encoding vector $E$ and a reconstruction vector $R$, so that the reconstruction $\hat{x}=R\cdot y$ of the encoding $y=E\cdot x$ is minimal using the L2-loss, ie.  $\min_{R,E} ||x-R\cdot E\cdot x||^2$.
%\begin{small}$$  \end{small}
The optimal solution which minimizes the reconstruction loss stems from projecting onto the eigenvector space (as given by a Principal Component Analysis)\cite{bal89}. That is, given there is just a single latent variable, the optimal solution for $W=R\cdot E$ is the first eigenvector $u_1$. This is illustrated in Figure \ref{fig:the} in the upper part with $y=u_1\cdot x$. For \emph{ClaDec} the goal is to explain a linear regression model $y=E\cdot x$. The vector $E$ is found by solving a regression problem. We fit the decoder $R$ to minimize the reconstruction loss on the original inputs given the encoding, ie.  $\min_{R} ||x-R\cdot y||^2$ with $y=E\cdot x$. The more similar the regression problem is to the encoding problem of an AE, the more similar are the reconstructions. Put differently, the closer $E$ is to $u_1$ the lower the reconstruction loss and the more similar are the optimal reconstructions for the reference AE and \emph{ClaDec}. Assume that $E$ differs strongly from $u_1$, ie. say that the optimal solution to the regression problem is the second eigenvector $y=u_2\cdot x$. This is shown in the lower part of Figure \ref{fig:the}. When comparing the optimal reconstruction of the \emph{RefAE}, ie. using $y=u_1x$, and the illustrated reconstruction of \emph{ClaDec}, ie. using $y=u_2x$, it becomes apparent that for the optimal encoding $y=u_1x$ the reconstructions of both coordinates $x_1$ and $x_2$ are fairly accurate on average. In contrast, using $y=u_2x$, coordinate $x_2$ is reconstructed more accurately (on average), whereas the reconstruction of $x_1$ is mostly very poor. 

Generally, this suggests that a representation obtained from a model (trained for some task) captures less information than an encoder optimized towards reconstructing inputs. But aspects of inputs relevant to the task should be captured relatively in more detail than those that are irrelevant. Reconstructions from \emph{ClaDec} should show more similarity to original inputs for attributes relevant to classification and less similarity for irrelevant attributes. But, overall reconstructions from the classifier will show less similarity to inputs than those of an AE.
% With reference to the proposed classifier-decoder architecture, this suggests that we expect

%that a classifier may capture some input attributes more accurately than the reference AE but overall the \emph{RefAE} provides a more accurate encoding.

%The introduction discussed some desirable properties of explanations, namely 
% While there are more properties such as fairness or privacy, the chosen ones are among the most crucial metrics that are also commonly dealt with in the literature.
\section{Assessing Interpretability and Fidelity} \label{sec:san} % and Their Measurement
%We focus on interpretability and fidelity. %We state objective, quantifiable, necessary conditions that explainability methods must achieve to hold these properties. %While there are many more properties and, ultimately, explanations must be judged by humans,  %The first condition relates to fidelity. It is of general nature, ie. it should be fulfilled by (any) explanation method. The second condition deals with interpretability. It is more tailored towards methods that provide explanations by synthesizing inputs.% We also briefly discuss robustness -- a property related to fidelity and interpretability.
%However, if inputs strongly impact encoding (even though they are similar) 
\textbf{Fidelity} is the degree to which an explanation captures model behavior. That is, a ``fidel'' explanation captures the decision process of the model accurately. The proposed evaluation (also serving as sanity check) uses the rational that fidel explanations for decisions of a well-performing model should be helpful in performing the task the model addresses. Concretely, training a new classifier $C^E_{eval}$ on explanations and, possibly, inputs should yield a better performing classifier than relying on inputs only. That is, we train a baseline classifier $C^R_{eval}(\hat{X}_R)$ on the reconstructions of the \emph{RefAE} and a second classifier with identical architecture $C^E_{eval}(\hat{X}_E)$ on explanations from \emph{ClaDec}. The latter classifier should achieve higher accuracy. This is a much stronger requirement than the common sanity check demanding that explanations must be valuable to perform a task better than a ``guessing'' baseline.  One must be careful that explanations do not contain additional external knowledge (not present in the inputs or training data) that help in performing the task. For most methods, including ours, this holds true. Therefore, it is not obvious that training on explanations allows to improve on classification performance compared to training on inputs that are more accurate reconstructions of the original inputs. Improvements seem only possible if an explanation is a more adequate representation to solve the problem. Formally, we measure the similarity between the reconstructions $\hat{X}_R$ (using \emph{RefAE}) and $\hat{X}_E$ (of \emph{ClaDec}) with the original inputs $X$. We show that explanations (from \emph{ClaDec}) bear less similarity with original inputs than reconstructions from \emph{RefAE}. Still, training on explanations $\hat{X}_E$ yields classifiers with better performance than training on the more informative outputs $\hat{X}_R$ from \emph{RefAE}.
    %This idea has been expressed by fur18
    %explanations are only computed using the model, the input and potentially the entire training data. This means that explanations are guaranteed to exclude information apart from the training data (and the input).
    
%(and Effort)  Effort is the cognitive load that is required by a human to make sense of an explanation. The assumption that familiarity results in reduced cognitive effort is well-justified. If unknown concepts are used in explanations then these novel concepts require additional explanations themselves, which are not needed for familiar concepts. In addition, Studies have shown that ``familiarity breeds trust'' \cite{gul95,che07}.
\noindent
\textbf{Interpretability}
is the degree to which the explanation is human understandable. We build upon the intuitive assumption that a human can better and more easily interpret explanations made of concepts that she is more familiar with. We argue that a user is more familiar with real-world phenomena and concepts as captured in the training data than possibly unknown concepts captured in representations of a NN. This implies that explanations that are more similar to the training data are more interpretable than those with strong deviation from the training data. Therefore, we quantify interpretability by measuring the distance to the original input, ie. the reconstruction loss. If explanations show concepts that are highly fidelitous, but non-intuitive for a user (high reconstruction loss) a user can experience difficulties in making sense of the explanation. In contrast, a trivial explanation (showing the unmodified input) is easy to understand but it will not reveal any insights into the model behavior, i.e., it lacks fidelity. %Furthermore, parts of the input are reconstructed in detail, if (i) they are identical across all images and if (ii) they are important for classification. In case parts of the input are not very relevant for classification they will be reconstructred in a more blurry manner.

\section{Evaluation}
\begin{figure}
  \centering
  \vspace{-12pt}
  \includegraphics[width=0.9\linewidth]{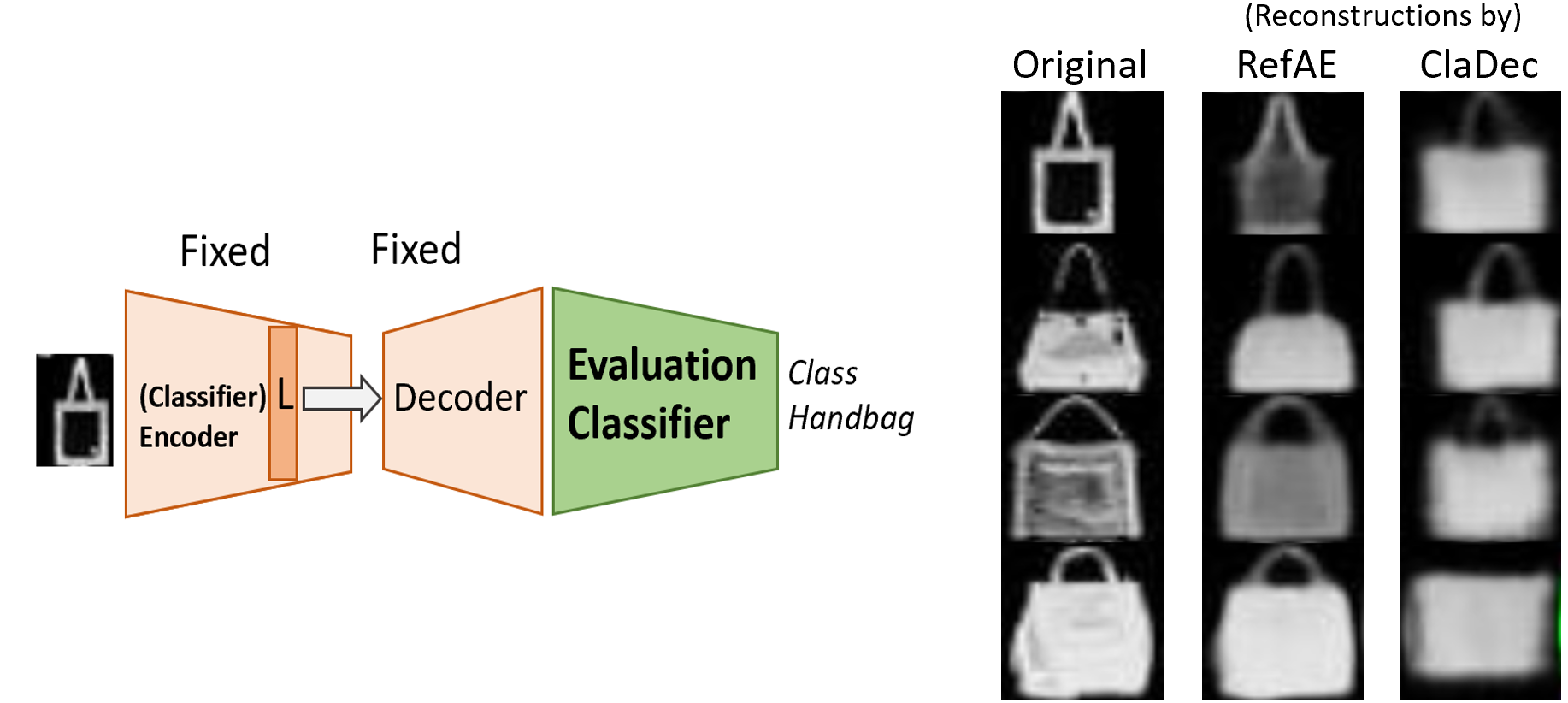} %in Table  \ref{tab:arch} 
  \vspace{-12pt}
  \caption{Left panel: Evaluation setup using a dedicated evaluation classifier. Right panel: Comparison of original inputs and reconstructions using the FC layer of the encoder for handbags. Comparing \emph{RefAE} and \emph{ClaDec} shows that both do not reconstruct detailed textures. The classifier does not rely on graytones, which are captured by \emph{RefAE}. It uses prototypical shapes.} \label{fig:recFC} %also with green indicating larger values of
  \vspace{-12pt}%\emph{RefAE} and red of the classifier
\end{figure}

In our qualitative and quantitative evaluation we focus on image classification using CNNs and the following experiments: (i) Explaining different layers for correct and incorrect classifications, (ii) Varying the fidelity and interpretability tradeoff.
Our decoder follows a standard design, ie. using 5x5 deconvolutional layers. For the classifier (and encoder) we used the same architecture, ie. a VGG-5 and ResNet-10. For ResNet-10 we reconstructed after each block. Both architectures behaved similiarly, thus we only report for VGG-5. Note, that the same classifier architecture (but trained with different input data) serves as encoder in \emph{RefAE}, classifier in \emph{ClaDec} and for classifiers used for evaluation of reconstructions, ie. classifier $C^E_{Eval}$ (for assessing \emph{ClaDec}) and  $C^R_{Eval}$ (for \emph{RefAE}). The evaluation setup is shown in the right panel of Figure \ref{fig:recFC} for \emph{ClaDec}. Thus, we denote by ``Acc Enc \emph{ClaDec}'' the validation accuracy of the encoder, ie. classifier, of the \emph{ClaDec} architecture and by ``Acc Eval \emph{RefAE}'' the validation accuracy of the classifier $C^R_{Eval}$ used for evaluation as shown in Figure \ref{fig:recFC} trained on reconstructions from the reference AE. Other combinations are analogous.
%We only report on validation accuracies, since training accuracies were above 99.5 percent.\footnote{Except for the benchmark, where we trained for less epochs}

%\cite{stan18}\cite{xia17}, MNIST and TinyImageNet
Note that the decoder architecture varies depending on which layer is to be explained. The original architecture allows to either obtain reconstructions from the last convolutional layer or the fully connected layer. For a lower layer, the highest deconvolutional layers from the decoder have to be removed, so that the reconstructed image $\hat{X}$ has the same width and height as the original input $X$. We employed three datasets namely Fashion-MNIST, MNIST and TinyImageNet. Since all datasets behaved similarly, we focus on Fashion-MNIST consisting of 70000 28x28 images of clothing stemming from 10 classes that we scaled to 32x32. 10000 samples are used for testing. We train all models using the Adam optimizer for 64 epochs. That is, the \emph{refAE}, the decoder of \emph{ClaDec}, the classifier serving as encoder in \emph{ClaDec} as well as the classifiers used for evaluation. We conducted 5 runs for each reported number. We show both averages and standard deviations.

% in Section \ref{sec:evalMtiny}.% For MNIST, we used the cross-entropy loss instead of the L2-loss.\footnote{Otherwise results were often of poor quality.} 

%  We advocate the use of Fashion MNIST over the more commonly used MNIST, since it is more difficult to obtain large accuracies (than MNIST) and images show more diversity, eg. images exhibit more diverse shapes and a richer set of grayscale tones. Still, for comparability we also show reconstructions of MNIST in the Appendix. 

%We provide also samples of MNIST in the Appendix. %We employ three datasets namely Fashion MNIST\cite{xia17}, MNIST and TinyImageNet\footnote{\url{https://tiny-imagenet.herokuapp.com/}}. We use MNIST for comparability, since it is still commonly used in explainability tasks. We advocate the use of Fashion MNIST since it is more difficult to obtain large accuracies (than MNIST) and images show more diversity, eg. images exhibit more diverse shapes and a richer set of grayscale tones rather than just binary pixels as for MNIST.
%  in its extended version of 9.

\begin{figure}
  \centering
  \vspace{-6pt}
  \includegraphics[width=1.0\linewidth]{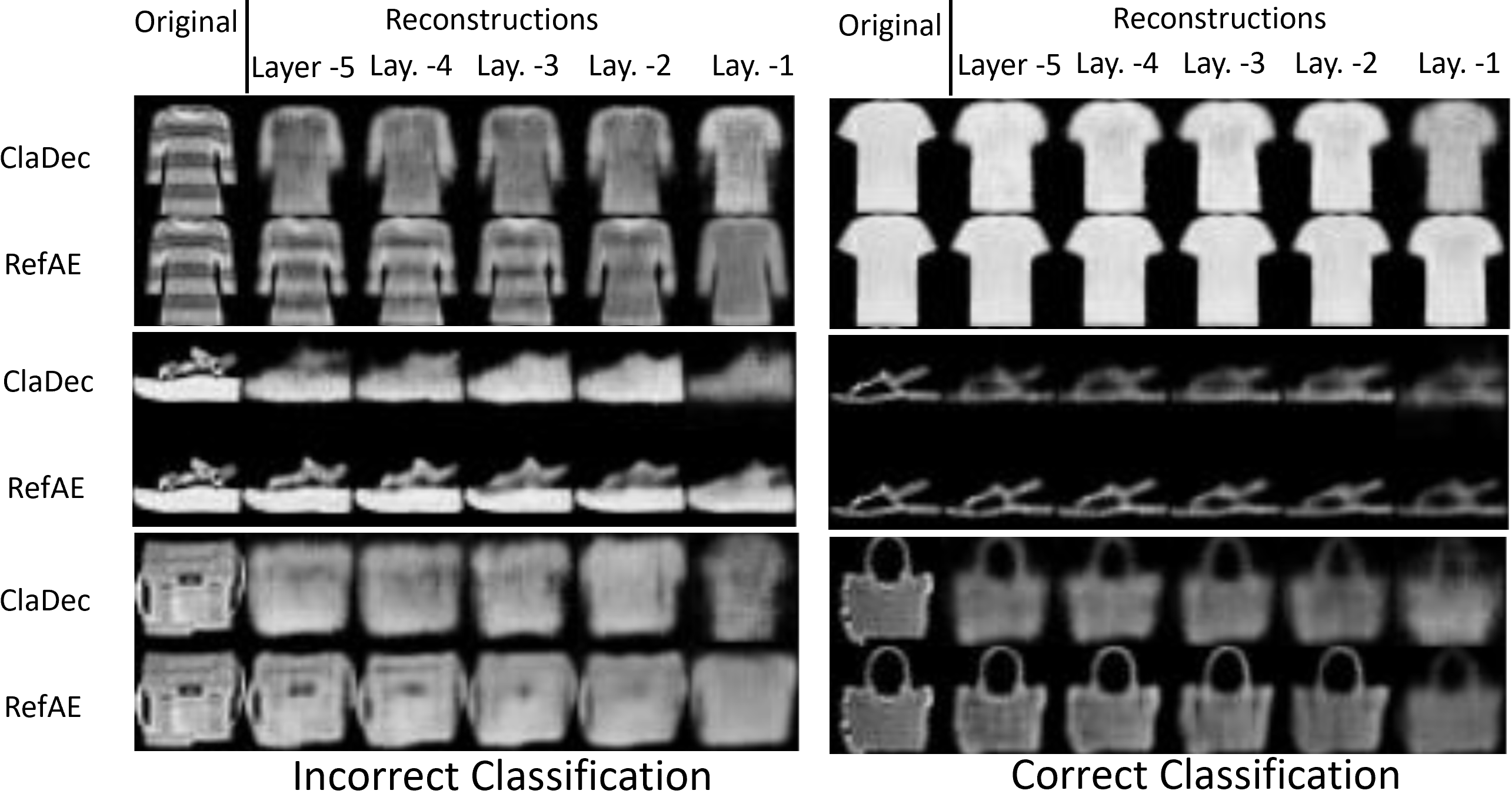} %\textbf{}in Table  \ref{tab:arch}
  \vspace{-12pt}
  \caption{Comparison of original inputs and reconstructions using multiple layers of the encoder. For incorrect samples it shows a gradual transformation into another class. Differences between \emph{RefAE} and \emph{ClaDec} increase with each layer \label{fig:cowr}} %also with green indicating larger values of \emph{RefAE} and red of the classifier
  \vspace{-6pt}
\end{figure}

\subsection{Qualitative Evaluation} \label{sec:qual}

% \begin{figure}
%   \centering
%   \includegraphics[width=\linewidth]{show_Comp-2.jpg}
%   \caption{Comparison of original and reconstructions using the last conv. layer of the encoder in Table  \ref{tab:arch}. Differences between reconstructions are shown in the last column.} \label{fig:show2}
% \end{figure}

% Overall, \emph{RefAE} is able to reconstruct shapes and graytones better. Comparing reconstructions from \emph{ClaDec} to those of \emph{RefAE} and the original inputs, it becomes apparent that reconstructions from \emph{ClaDec} have more uniform graytones, ie. they are poor approximations of the actual graytones compared to both the original and reconstructions from \emph{RefAE}.%(see Table \ref{tab:arch})

\noindent\textbf{Varying Explanation Layers}: Reconstructions based on \emph{RefAE} and \emph{ClaDec} are shown in  Figures \ref{fig:recFC} and \ref{fig:cowr}. For the last layer, ie. the fully connected (FC) layer, there is only one value per class, implying a representation of 10 dimensions for Fashion-MNIST. For the handbags depicted in Figure \ref{fig:recFC} and explained in the caption, comparing the original inputs and the reconstructions by \emph{RefAE} and \emph{ClaDec} shows clear differences in reconstructions. Some conclusions are knowledge of precise graytones is not used to classify these objects. Reconstructions from \emph{ClaDec} resemble more prototypical, abstract features of handbags. Figure \ref{fig:cowr} shows reconstructions across layers. For all samples one can observe a gradual abstraction resulting in change of shape and graytones as well as loss of details. The degree of abstraction varies significantly among samples, e.g. modest for T-Shirt and strong for handbags in the right panel. Reconstructions from \emph{ClaDec} are more blurry than for \emph{RefAE}. Blurriness indicates that the representation of the layer does not contain information needed to recover the details. However, the reason is not (primarily) distortions inherent in the decoder architecture, since \emph{RefAE} produces significantly sharper images, but rather the abstraction process of the classifier. This is most apparent for incorrectly classified samples (left panel). One can observe a gradual modification of the sample into another class. This helps in understanding, how the network changed input features and at which layer. For example, the sandal (second row) appears like a sneaker at layer -3, whereas the reconstruction from \emph{RefAE} still maintains the look of a sandal. The black ``holes'' in the sandals have vanished at layer -3 for the incorrectly classified sandal, whereas for the correctly classified sandal (right panel, same row), these holes remain. The bag (third row) only shows signs of a T-shirt in the second layer, where stumps of arms appear.
\begin{figure}
\vspace{-6pt}
\centering \includegraphics[width=0.9\linewidth]{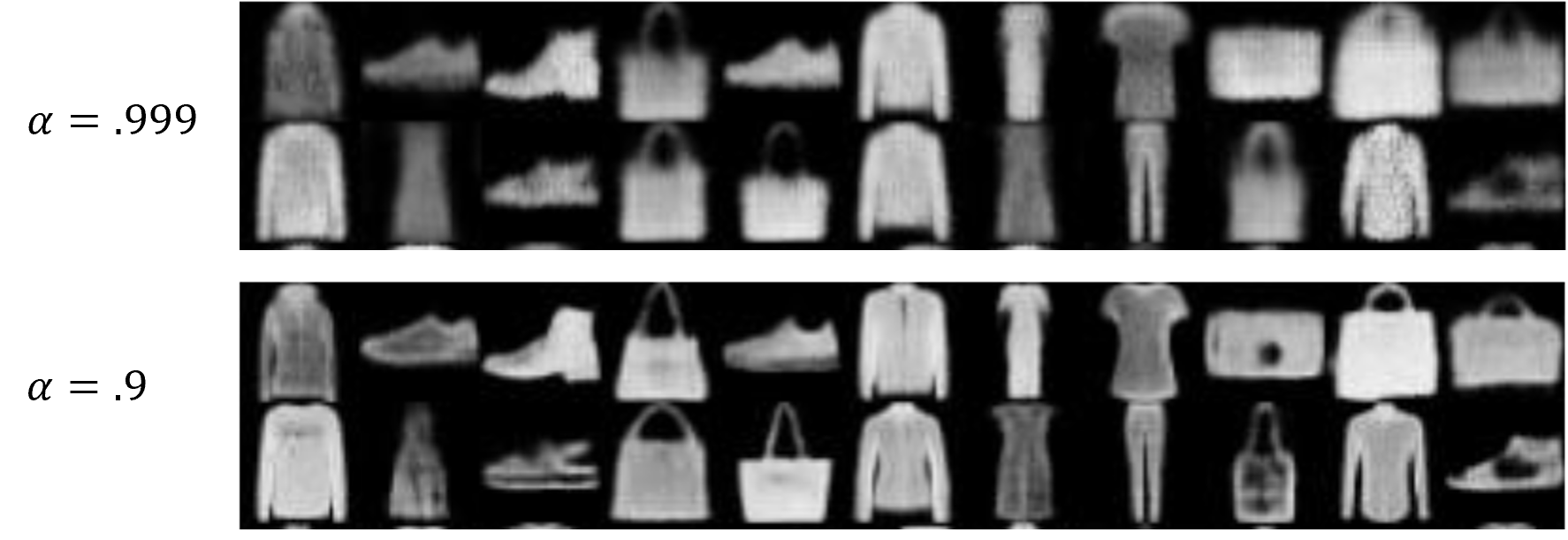}
\vspace{-12pt}
  \caption{Adding classification loss ($\alpha>0$) yields worse reconstructions for the last conv. layer. Using classification loss only, reconstructions are not human recognizable.} \label{fig:trade}
  \vspace{-6pt}
\end{figure}

\noindent\textbf{Fidelity and Interpretability Trade-off}: Figure \ref{fig:trade} shows for the last conv. layer (second to last overall) the impact of adding a classification loss (Figure \ref{fig:deco}) to modulate how much the model impacts reconstructions. Neglecting reconstruction loss, i.e. $\alpha=1$, yields non-surprisingly non-interpretable reconstructions (not shown in Figure). Already modest reconstruction loss leads to well-recognizable shapes. The quality of reconstructions in terms of sharpness and amount of captured detail constantly improves the more emphasis is put on reconstruction loss. It also becomes evident that the NN learns ``prototypical'' samples (or features) towards which reconstructed samples are being optimized. For example, the shape of handbag handles shows much more diversity for values of $\alpha$ close to 0, it is fairly uniform for relatively large values of $\alpha$. %It is also interesting to note that while most details of images vanish, when reducing $\alpha$, others become more profound or remain, eg. the gap between legs of pants.
Thus, the parameter $\alpha$ provides a means to reconstruct a compromise between the sample that yields minimal classification loss and a sample that is true to the input. It suggests that areas of the reconstruction of \emph{ClaDec} that are similar to the original input are also similar to a ``prototype'' that minimizes classification loss. That is, the network can recognize them well, whereas areas that are strongly modified, resemble parts that seem non-aligned with ``the prototype'' encoded in the network.

\begin{table}
\footnotesize
\begin{tabular}{| l|l|l|| l|l|l|l|}\hline
Layer&Rec Loss \emph{ClaDec}&Rec Loss \emph{RefAE}&$\Delta$&Acc Eval \emph{ClaDec}&Acc Eval \emph{RefAE}&$\Delta$  \\ \hline
-1&28.6\text{\tiny{$\pm$0.6851}}&8.48\text{\tiny{$\pm$0.3799}}&20.2&0.89\text{\tiny{$\pm$0.0031}}&0.83\text{\tiny{$\pm$0.0105}}&0.06\\ \hline
-3&4.56\text{\tiny{$\pm$0.0921}}&3.63\text{\tiny{$\pm$0.0729}}&0.93&0.877\text{\tiny{$\pm$0.0074}}&0.863\text{\tiny{$\pm$0.0093}}&0.014\\ \hline
-5&1.93\text{\tiny{$\pm$0.1743}}&1.87\text{\tiny{$\pm$0.0933}}&0.06&0.878\text{\tiny{$\pm$0.0073}}&0.875\text{\tiny{$\pm$0.0048}}&0.003\\ \hline
\end{tabular}
\caption{Explaining layers: \emph{ClaDec} has larger reconstruction loss but the evaluation classifier has higher accuracy on \emph{ClaDec}'s reconstructions } \label{tab:lay}
\vspace{-18pt}
\end{table}

\begin{table}
\footnotesize
\begin{tabular}{|l| l| l|l| l|}\hline
$\alpha$&Total Loss  \emph{ClaDec} &Rec Loss& Classifier Loss &Acc Eval \emph{ClaDec}  \\ \hline
.0&0.01\text{\tiny{$\pm$0.0033}}&285.5\text{\tiny{$\pm$52.01}}&0.0\text{\tiny{$\pm$0.0}}&0.9028\text{\tiny{$\pm$0.0035}}\\ \hline
.001&0.03\text{\tiny{$\pm$0.0023}}&25.4\text{\tiny{$\pm$0.9292}}&0.03\text{\tiny{$\pm$0.0009}}&0.9033\text{\tiny{$\pm$0.0022}}\\ \hline
%0.01&0.138\text{\tiny{$\pm$0.0062}}&13.2\text{\tiny{$\pm$0.5637}}&-0.1305\text{\tiny{$\pm$0.0056}}&0.903\text{\tiny{$\pm$0.0019}}\\ \hline
.1&0.84\text{\tiny{$\pm$0.0132}}&8.35\text{\tiny{$\pm$0.1195}}&0.75\text{\tiny{$\pm$0.0106}}&0.9011\text{\tiny{$\pm$0.0026}}\\ \hline
1.0&7.49\text{\tiny{$\pm$0.1119}}&7.49\text{\tiny{$\pm$0.1119}}&4.4\text{\tiny{$\pm$0.254}}&0.8824\text{\tiny{$\pm$0.0042}}\\ \hline
 \end{tabular}
\caption{Adding classification loss $\alpha>0$ (Equation \ref{eq:alpha}) yields worse reconstructions, but higher evaluation accuracy}\label{tab:acc}
\vspace{-18pt}
\end{table}

\vspace{-6pt}
\subsection{Quantitative Evaluation} \label{sec:quant}
\vspace{-6pt}
\noindent\textbf{Varying Explanation Layers}: Results in Table \ref{tab:lay} contain two key messages: First, the reconstruction loss is lower for \emph{RefAE} than for \emph{ClaDec}. This is expected since \emph{RefAE} is optimized entirely towards minimal reconstruction loss of the original inputs. Second, the classification (evaluation) accuracy is higher, when training the evaluation classifier $C_{Eval}$ using reconstructions from \emph{ClaDec} than from \emph{RefAE}. This behavior is not obvious, since the reconstructions from \emph{ClaDec} are poorer according to the reconstruction loss. That is, they contain less information about the original input than those from \emph{RefAE}. However, it seems that the ``right'' information is encoded using a better suited representation. Aside from these two key observations there are a set of other noteworthy behaviors: As expected the reconstruction loss increases the more encoder layers, ie. the more transformations of the input, are used. The impact is significantly stronger for \emph{ClaDec}. The difference between \emph{RefAE} and \emph{ClaDec} increases the closer the layer to explain is to the output. This is not surprising, since lower layers are known to be fairly general, ie. in transfer learning lower layers are the most applicable to work well for varying input data. There is a strong increase for the last layer, this is also no surprise, since the last layer consists of fairly fewer dimensions, i.e. 10 dimensions (one per class) compared to more than 100 for the second last layer. The classification accuracy for the evaluation classifier somewhat improves the more layers are used as encoder, ie. of the classifier that should be explained. The opposite holds for \emph{RefAE}. This confirms that \emph{RefAE} focuses on the wrong information, whereas the classifier trained towards the task focuses on the right information and encodes it well.\\
\noindent\textbf{Fidelity and Interpretability Tradeoff}: Table \ref{tab:acc} shows that evaluation accuracy increases when adding a classification loss, i.e. $\alpha>0$ yields an accuracy above 90\% whereas $\alpha=0$ gives about 88\%. Reconstructions that are stronger influenced by the model to explain (larger $\alpha$) are more truthful to the model, but they exhibit larger differences from the original inputs. Choosing $\alpha$ slightly above the minimum, i.e. larger than 0, already has a strong impact.

\vspace{-3pt}
\section{Related Work}% as claimed in \cite{adebayo2018sanity,kin19,gho19} %adadi2018peeking,%We discuss approaches to visualize single features as well as to understand particular decisions.
\vspace{-3pt}
We categorize explainability methods~\cite{schneider2019pers} into methods that synthesize inputs (like ours and \cite{ngu16,yos15}) and methods that rely on saliency maps\cite{sim13} based on perturbation~\cite{ribeiro2016should,zeil14} or gradients~\cite{selvaraju2017grad,bach2015pixel}. Saliency maps show feature importance of inputs, whereas synthesized inputs often show higher level representations encoded in the network. Perturbation-based methods include occlusion of parts of the inputs \cite{zeil14} and investigating the impact on output probabilities of specific classes. Linear proxy models such as LIME~\cite{ribeiro2016should} perform local approximations of a black-box model using simple linear models by also assessing modified inputs. Saliency maps\cite{sim13} highlight parts of the inputs that contributed to the decision. Many explainability methods have been under scrutiny for failing sanity checks \cite{adebayo2018sanity} and being sensitive to factors not contributing to model predictions \cite{kin19} or adversarial perturbations \cite{gho19}. We anticipate that our work is less sensitive to targeted, hard to notice perturbations \cite{gho19} as well as translations or factors not impacting decisions \cite{kin19}, since we rely on encodings of the classifier. Thus, explanations only change if these encodings change, which they should. The idea to evaluate explanations on downstream tasks is not new, however a comparison to a ``close'' baseline like our \emph{RefAE} is. Our ``evaluation classifier'' using only explanations (without inputs) is more suitable than methods like \cite{sch20ref} that use explanations together with inputs in a more complex, non-standard classification process. Using inputs and explanations for the evaluation classifier is diminishing differences in evaluation outcomes for any compared methods since a network might take missing information in the explanation from the input. So far, inputs have only been synthesized to understand individual neurons through activation maximization in an optimization procedure\cite{ngu16}. The idea is to identify inputs that maximize the activation of a given neuron. This is similar to the idea to identify samples in the input that maximize neuron activation. \cite{ngu16} uses a (pre-trained) GAN on natural images relevant to the classification problem. It identifies through optimization the latent code that when fed into the GAN results in a more or less realistic looking image that maximally activates a neuron \cite{yos15} uses regularized optimization as well, yielding artistically more interesting but less recognizable images. Regularized optimization has also been employed in other forms of explanations of images, e.g. to make human understand how they can alter visual inputs such as handwriting for better recognizability by a CNN \cite{sch20hu}. %Our idea to verify if a ``distortion'' of realism is due to the decoder might also be beneficial to improve on explanations of \cite{ngu16}. Our approach does not require any optimization per instance to be explained. %Note, that any approach that aims at explicitly interpreting individual neurons (or representations) such as \cite{ngu16}, cannot be easily extended to explain the entire model or layer behavior. It suffers from the problem that networks allowing to distinguish just few classes still have hundreds or even thousands of neurons (per layer). Thus, while explaining neurons is highly important, it is a conceptually different problem from explaining a decision or an entire layer.  %As highlighted in Figure \ref{fig:met} there are additional differences between our method and \cite{ngu16}. 
%Other ideas for explanations include \cite{kim17,shr17,koh2017,ghor19}. 
\cite{kim17,ghor19} allow to investigate  high level concepts that are relevant to a specific decision. DeepLift\cite{shr17} compares activations to a reference and propagates them backwards. Defining the reference is non-trivial and domain specific. \cite{koh2017} estimates the impact of individual training samples. %\cite{liu19tow} discusses how to explain variational AEs using gradient-based methods. %We also propose a method to explain AEs or, more precisely, just the impact of training an encoder in an AE architecture in Section \ref{sec:expAE}. \paragraph{Auto-encoders} \noindent\emph{Auto-encoders}:
\cite{van20} uses a variational AE for contrastive explanations. They use distances in latent space to identify samples which are closest to a sample $X$ of class $Y$ but actually classified as $Y'$.
%Given that one should explain why a sample $X$ is of class $Y$ and not of $Y'$, they compute the latent representation of $X$. Find the closest sample $X'$ in latent space of class $Y'$. They sample points between $X$ and $X'$ in latent space and reconstruct images based on the code provided by the sampled points. They return the image being classified as class $Y'$ with latent code closest to $X$.% From our perspective it can be valuable to replace the latent representation produced by the VAE, with a representation of the classifier.
%Denoising AEs are well established \cite{vin10,du16}. They can be used to remove noise from images, reconstruct images and leverage data in an unsupervised manner \cite{vin10,du16}. Ideas to combine unsupervised learning approaches to remove noise and supervised learning by extending loss functions have been presented since the early 90's\cite{dec93}. In the context of explanations, \cite{qi19co} used an AE with skip-connections for saliency map predictions.

%Use AE and add classifier based on latent rep. \cite{li2020}.

%\medskip

\section{Conclusions}

%\noindent\textbf{Conclusions}\\ % such as sound theoretical grounding and evaluation. We have proposed a theoretically grounded  Explaining complex deep learning models is difficult as witnessed by multiple works pointing out shortcomings of many existing explanation methods.
Our explanation method synthesizes human understandable inputs based on layer activations. It takes into account distortions originating from the reconstruction process. It is verified using novel sanity checks. In the future, we plan to to investigate differences among networks, eg. a result of model fine-tuning as in personalization \cite{sch20pers} or to look at subsets of layer activations. %We believe that our method might form the basis for many more methods that further expand and contribute to the field of explainability. %In the future, we aim to investigate other types of data as well as additional explanation properties.

%\section{Acknowledgments}
%We thank Jeroen van Doorenmalen for valuable discussions.

%\bibliographystyle{APA}

\bibliographystyle{splncs04}
\bibliography{refs}

\begin{thebibliography}{10}
\providecommand{\url}[1]{\texttt{#1}}
\providecommand{\urlprefix}{URL }
\providecommand{\doi}[1]{https://doi.org/#1}

\bibitem{adebayo2018sanity}
Adebayo, J., Gilmer, J., Muelly, M., Goodfellow, I., Hardt, M., Kim, B.: Sanity
  checks for saliency maps. In: Neural Information Processing Systems (2018)

\bibitem{bach2015pixel}
Bach, S., Binder, A., Montavon, G., Klauschen, F., M{\"u}ller, K.R., Samek, W.:
  On pixel-wise explanations for non-linear classifier decisions by layer-wise
  relevance propagation. PloS one  \textbf{10}(7) (2015)

\bibitem{bal89}
Baldi, P., Hornik, K.: Neural networks and principal component analysis:
  Learning from examples without local minima. Neural networks  \textbf{2}(1),
  53--58 (1989)

\bibitem{van20}
van Doorenmalen, J., Menkovski, V.: Evaluation of cnn performance in
  semantically relevant latent spaces. In: Int. Symposium on Intelligent Data
  Analysis (2020)

\bibitem{gho19}
Ghorbani, A., Abid, A., Zou, J.: Interpretation of neural networks is fragile.
  In: AAAI Conference on Artificial Intelligence (2019)

\bibitem{ghor19}
Ghorbani, A., Wexler, J., Zou, J.Y., Kim, B.: Towards automatic concept-based
  explanations. In: Advances in Neural Information Processing Systems (2019)

\bibitem{kim17}
Kim, B., Wattenberg, M., Gilmer, J., Cai, C., Wexler, J., Viegas, F., Sayres,
  R.: Interpretability beyond feature attribution: Quantitative testing with
  concept activation vectors (tcav). arXiv preprint arXiv:1711.11279  (2017)

\bibitem{kin19}
Kindermans, P.J., Hooker, S., Adebayo, J., Alber, M., Sch{\"u}tt, K.T.,
  D{\"a}hne, S., Erhan, D., Kim, B.: The (un) reliability of saliency methods.
  In: Explainable AI: Interpreting, Explaining and Visualizing Deep Learning
  (2019)

\bibitem{koh2017}
Koh, P.W., Liang, P.: Understanding black-box predictions via influence
  functions. In: Proc. of Int. Conference on Machine Learning (2017)

\bibitem{ngu16}
Nguyen, A., Dosovitskiy, A., Yosinski, J., Brox, T., Clune, J.: Synthesizing
  the preferred inputs for neurons in neural networks via deep generator
  networks. In: Advances in neural information processing systems. pp.
  3387--3395 (2016)

\bibitem{ribeiro2016should}
Ribeiro, M.T., Singh, S., Guestrin, C.: Why should i trust you?: Explaining the
  predictions of any classifier. In: SIGKDD (2016)

\bibitem{sch20hu}
Schneider, J.: Human-to-ai coach: Improving human inputs to ai systems. In:
  International Symposium on Intelligent Data Analysis (2020)

\bibitem{schneider2019pers}
Schneider, J., Handali, J.P.: Personalized explanation for machine learning: a
  conceptualization. In: European Conference on Information Systems (ECIS)
  (2019)

\bibitem{sch20ref}
Schneider, J., Vlachos, M.: Reflective-net: Learning from explanations. In:
  arxiv: 2011.13986 (2020)

\bibitem{sch20pers}
Schneider, J., Vlachos, M.: Personalization of deep learning. In: Data
  Science--Analytics and Applications. Springer (2020)

\bibitem{selvaraju2017grad}
Selvaraju, R.R., Cogswell, M., Das, A., Vedantam, R., Parikh, D., Batra, D.:
  Grad-cam: Visual explanations from deep networks via gradient-based
  localization. In: IEEE International Conference on Computer Vision (ICCV).
  pp. 618--626 (2017)

\bibitem{shr17}
Shrikumar, A., Greenside, P., Kundaje, A.: Learning important features through
  propagating activation differences. In: Int. Conf. on Machine Learning (2017)

\bibitem{sim13}
Simonyan, K., Vedaldi, A., Zisserman, A.: Deep inside convolutional networks:
  Visualising image classification models and saliency maps. arXiv preprint
  arXiv:1312.6034  (2013)

\bibitem{yang19}
Yang, F., Du, M., Hu, X.: Evaluating explanation without ground truth in
  interpretable machine learning. arXiv preprint arXiv:1907.06831  (2019)

\bibitem{yos15}
Yosinski, J., Clune, J., Nguyen, A., Fuchs, T., Lipson, H.: Understanding
  neural networks through deep visualization. arXiv preprint arXiv:1506.06579
  (2015)

\bibitem{zeil14}
Zeiler, M.D., Fergus, R.: Visualizing and understanding convolutional networks.
  In: European conference on computer vision (2014)

\end{thebibliography}

%For journal
% \begin{figure}[h]
%  %   \vspace{-6pt}
%   \centering
%   \includegraphics[width=0.9\linewidth]{saliency.PNG}
%   \caption{Saliency maps examples. Taken from \cite{yeh19}.} \label{fig:sal}
% %  \vspace{-15pt}
% \end{figure}

\end{document}

\title{Supplementary material \\ Explaining Neural Networks by Decoding Layer Activations} 
%\author{\IEEEauthorblockN{Johannes Schneider}\IEEEauthorblockA{Institute of Information Systems, \\University of Liechtenstein \\Vaduz,Liechtenstein \\johannes.schneider@uni.li}}
\maketitle %
\begin{abstract} 
This document contains supplementary material, which is referenced in the main document.
\end{abstract}

\section{Architectures}

For TinyImageNet, we doubled the number of conv layers  of the encoder in Table \ref{tab:arch} and increased the number of neurons of all layers (encoder and decoder) by a factor of four.
 \begin{table}[h] 	
 	\vspace{-6pt}
 	\begin{center}
 		\scriptsize
 		%\footnotesize
 		\setlength\tabcolsep{2.5pt}
% 		\rotatebox{90}{
	%\begin{minipage}{.49\linewidth}
	\centering
 		\begin{tabular}{| l | l| l|l| }\hline
 			\multicolumn{2}{|c|}{VGG-style Encoder} & \multicolumn{2}{c|}{Decoder}\\  \hline
			Type/Stride & Filter Shape &Type/Stride& Filter Shape \\  \hline
 			C/s2     & $3\tiny{\times} 3 \tiny{\times} 1 \tiny{\times} 16$ & FC     & nClasses  \\ \hline
 			C/s2     & $3\tiny{\times} 3 \tiny{\times} 16 \tiny{\times} 32$ & DC/s2     & $3\tiny{\times} 5 \tiny{\times} 5 \tiny{\times} 256$  \\ \hline
 			C/s2     & $3\tiny{\times} 3 \tiny{\times} 32 \tiny{\times} 64$ & DC/s2     & $3\tiny{\times} 5 \tiny{\times} 5 \tiny{\times} 128$  \\ \hline
 			C/s2     & $3\tiny{\times} 3 \tiny{\times} 64 \tiny{\times} 128$ & DC/s2     & $3\tiny{\times} 5 \tiny{\times} 5 \tiny{\times} 64$  \\ \hline
 			C/s2     & $3\tiny{\times} 3 \tiny{\times} 128 \tiny{\times} 256$ & DC/s2     & $3\tiny{\times} 5 \tiny{\times} 5 \tiny{\times} 32$  \\ \hline
 			FC/s1 & $256 \tiny{\times} $nClasses &     DC/s2 & $3\tiny{\times} 5 \tiny{\times} 5 \tiny{\times} 1$  \\ \hline
 			Softmax/s1 & Classifier &   &  \\ \hline
 			\end{tabular}
` 	%	\end{minipage}
 	\end{center}
 	\caption{Encoder/Decoder, where ``C'' is a convolution, ``DC'' a deconv; a BatchNorm and a ReLu layer follow each ``C'' layer; a ReLu layer follows each ``DC'' layer}  \label{tab:arch} 
 	\vspace{-6pt}
 \end{table}

\section{Additional Experiments}
We conducted two more experiments:
\subsubsection{Explaining AE and the Impact of Classifier Performance}\label{sec:expAE}
\begin{figure}
  \centering
  \includegraphics[width=\linewidth]{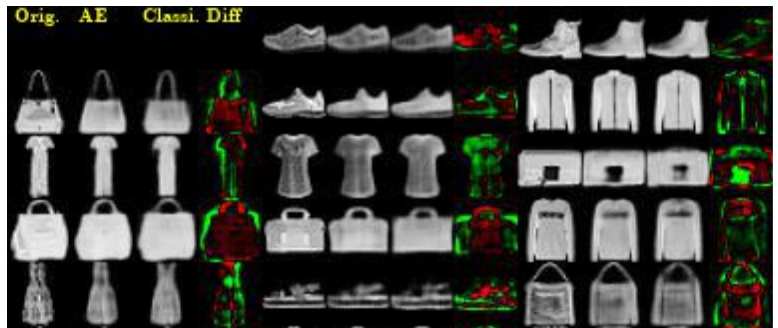}
  \caption{Comparison of original inputs and reconstructions using the last conv. layer of the encoder in Table  \ref{tab:arch} without any training of the classifier in \emph{ClaDec}.} \label{fig:showNoTrain}
\end{figure}
 While our work focuses on explaining classifiers, we also briefly discuss how to explain an AE architecture itself, ie. the encoder in AE architecture. Figure \ref{fig:showNoTrain} shows reconstructions if the classifier is not trained at all. Interestingly they exhibit fairly good quality. We explain this for the quantitative assessment. Furthermore, comparing reconstructions of the reference AE with those of \emph{ClaDec} for an untrained classifier might be used to assess the relevance of training the encoder of an AE itself, ie. ``How does training an encoder (of an AE architecture) impact reconstructions (compared to a random encoder)?'' Based on Figure \ref{fig:showNoTrain} one might conclude that a trained encoder does lead to encodings that allow to better reconstruct original inputs than when using an untrained encoder utilizing randomly initialized layers. While sharpness is generally comparable for both reconstructions, there are several examples, where shapes of objects are altered or some details are missing, if an encoder is not-trained.

\subsubsection{Impact of Classifier Performance}
Table \ref{tab:ep} shows for \emph{ClaDec} that classifiers that are trained longer and, therefore, achieve higher validation accuracy also lead to better accuracy for the evaluation classifier. While this is expected, the dependence of reconstruction loss on the number of training epochs is more intricate. It is lowest without any training, increases quickly and then steadily decreases again. This pattern is highly statistically significant, ie. we conducted t-tests to verify that means between subsequent rows are different, yielding p-values below 0.01. When taking a closer look, it is not so surprising that an untrained network, ie. using random weights, achieves lower reconstruction loss than the trained classifier. First, it should be noted that the reconstruction loss using random weights is significantly higher as for the reference architecture, where the encoder is optimized. Second, it is known from extreme learning, eg. \cite{sun2017}, that encoders with randomly chosen weights can yield good results, if just the decoder is optimized. More generally, this phenomena might be traced back to the behavior of random projections formulated in the Johnson–Lindenstrauss lemma, saying that random projections yield good dimensionality reduction properties. The theorem is commonly used for dimensionality reduction in many contexts. Training of the classifier, seems to destroy some of the desirable properties of random initialization by focusing on information needed for classification (but not for reconstruction) -- as motivated theoretically. The reconstruction improves with more training, indicating that the initial encodings are noisy. But the reconstruction loss seems to converge (with increased training) towards a higher loss than for random initialization. %, eg. \cite{sch13} (see Figure \ref{fig:the})
\begin{table*}[]
\begin{tabular}{|l|l||l|l|l|| l|l|l|}\hline
Training Epochs of &Acc. of Classifier & Rec. Loss & Rec. Loss &$\Delta$&Acc Eval &Acc Eval &$\Delta$  \\
Classifier (to be explained)  &(to be explained) & \emph{ClaDec} & \emph{RefAE}& &\emph{ClaDec} & \emph{RefAE}&  \\
\hline
%clepr&CL Va Acc &AE LossC&AE Loss A&$\Delta$&Cl AccC& Cl AccA&$\Delta$  \\ \hline
0 & 0.1\tiny{$\pm0.0$}&5.445\tiny{$\pm0.164$}&3.333\tiny{$\pm0.04$}&2.112\tiny{$\pm0.159$}&0.85\tiny{$\pm0.003$}&0.886\tiny{$\pm0.004$}&-0.036\tiny{$\pm0.005$} \\ \hline
1 & 0.506\tiny{$\pm0.012$}&6.417\tiny{$\pm0.152$}&3.3\tiny{$\pm0.038$}&3.118\tiny{$\pm0.156$}&0.88\tiny{$\pm0.002$}&0.888\tiny{$\pm0.003$}&-0.007\tiny{$\pm0.004$} \\ \hline
4 & 0.885\tiny{$\pm0.003$}&6.608\tiny{$\pm0.079$}&3.299\tiny{$\pm0.049$}&3.309\tiny{$\pm0.088$}&0.893\tiny{$\pm0.004$}&0.893\tiny{$\pm0.004$}&-0.0\tiny{$\pm0.005$} \\ \hline
16 & 0.902\tiny{$\pm0.003$}&6.233\tiny{$\pm0.145$}&3.334\tiny{$\pm0.062$}&2.898\tiny{$\pm0.116$}&0.896\tiny{$\pm0.005$}&0.891\tiny{$\pm0.003$}&0.005\tiny{$\pm0.005$} \\ \hline
64 & 0.904\tiny{$\pm0.003$}&6.081\tiny{$\pm0.069$}&3.341\tiny{$\pm0.062$}&2.74\tiny{$\pm0.097$}&0.895\tiny{$\pm0.002$}&0.889\tiny{$\pm0.004$}&0.006\tiny{$\pm0.004$} \\ \hline
\end{tabular}
\caption{Impact of Classifier Accuracy (modulated through training epochs): Evaluation Accuracy increases with higher classifier accuracy as expected, behavior of rec.loss follows an inverted U shape.}\label{tab:ep}
\vspace{-16pt}
\end{table*}

\section{Quantitative And Qualitative Evaluation for ResNet}

\begin{table*}[h]
\begin{tabular}{|l| l| l|l| l|}\hline
$\alpha$&Total Loss \emph{ClaDec} &Rec Loss& Classifier Loss &Acc Eval \emph{ClaDec}  \\ \hline
0.0&0.0091\text{\tiny{$\pm$0.0012}}&416.5\text{\tiny{$\pm$76.69}}&0.0\text{\tiny{$\pm$0.0}}&0.9117\text{\tiny{$\pm$0.0026}}\\ \hline
0.001&0.0425\text{\tiny{$\pm$0.0029}}&34.5\text{\tiny{$\pm$1.604}}&0.0345\text{\tiny{$\pm$0.0016}}&0.9107\text{\tiny{$\pm$0.0023}}\\ \hline
0.01&0.1949\text{\tiny{$\pm$0.0085}}&19.0\text{\tiny{$\pm$0.7042}}&0.188\text{\tiny{$\pm$0.007}}&0.911\text{\tiny{$\pm$0.0032}}\\ \hline
0.1&1.33\text{\tiny{$\pm$0.0168}}&13.3\text{\tiny{$\pm$0.1638}}&1.193\text{\tiny{$\pm$0.0147}}&0.9087\text{\tiny{$\pm$0.0019}}\\ \hline
1.0&12.2\text{\tiny{$\pm$0.1964}}&12.2\text{\tiny{$\pm$0.1964}}&4.936\text{\tiny{$\pm$0.0572}}&0.8982\text{\tiny{$\pm$0.0052}}\\ \hline
 \end{tabular}
\caption{For ResNet, adding classification loss $\alpha<1$  yields worse reconstructions, but higher evaluation accuracy}\label{tab:accRes} %(Equation \ref{eq:alpha})
\vspace{-12pt}
\end{table*}

\begin{table*}[h]
\begin{tabular}{| l|l|l|| l|l|l|l|}\hline
Layer&(Rec.) Loss \emph{ClaDec}&(Rec.) Loss \emph{RefAE}&$\Delta$&Acc Eval \emph{ClaDec}&Acc Eval \emph{RefAE}&$\Delta$  \\ \hline
-1&34.0\text{\tiny{$\pm$0.6387}}&9.24\text{\tiny{$\pm$0.3874}}&24.8\text{\tiny{$\pm$0.8683}}&0.908\text{\tiny{$\pm$0.0019}}&0.804\text{\tiny{$\pm$0.0145}}&0.104\text{\tiny{$\pm$0.0149}}\\ \hline
-3&2.79\text{\tiny{$\pm$0.3186}}&1.92\text{\tiny{$\pm$0.1176}}&0.867\text{\tiny{$\pm$0.4051}}&0.89\text{\tiny{$\pm$0.0038}}&0.894\text{\tiny{$\pm$0.0043}}&-0.004\text{\tiny{$\pm$0.0058}}\\ \hline
-5&1.43\text{\tiny{$\pm$0.1404}}&1.38\text{\tiny{$\pm$0.1798}}&0.055\text{\tiny{$\pm$0.222}}&0.893\text{\tiny{$\pm$0.0046}}&0.892\text{\tiny{$\pm$0.0029}}&0.0\text{\tiny{$\pm$0.0066}}\\ \hline
\end{tabular}
\caption{Explaining different layers for ResNet: \emph{ClaDec} has larger reconstruction loss but the evaluation classifier on reconstructions from \emph{ClaDec} achieves higher accuracy.} \label{tab:layRes}
\vspace{-6pt}
\end{table*}

%\section{Qualitative Evaluation}
We show some more examples for the last layer for VGG.

\begin{figure}[h]
  \centering
  \includegraphics[width=\linewidth]{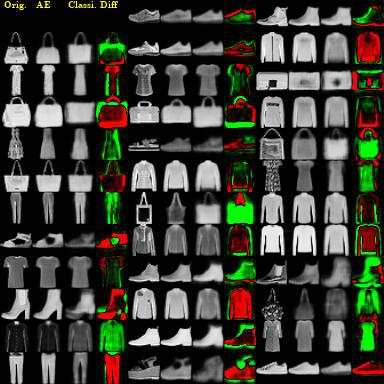}
  \caption{Comparison of original inputs and reconstructions using the last layer, ie. FC, of the encoder in Table  \ref{tab:arch}. Differences between reconstructions are shown in the last column.} \label{fig:show1} %also with green indicating larger values of \emph{RefAE} and red of the classifier
\end{figure}

Looking closer into multiple samples of handbags shows that handbags might be characterized by having a handle or not. Handbags without handles often have a rectangular shape. Figure \ref{fig:show1} shows that reconstructions capture this well: The reconstructed handbags that have a handle exhibit typically more of a square shape, whereas the handbags without handles are more of a rectangular shape. 

\section{Related Work - Comparison of methods}
\begin{figure*}[h]
  \centering
  \includegraphics[width=\linewidth]{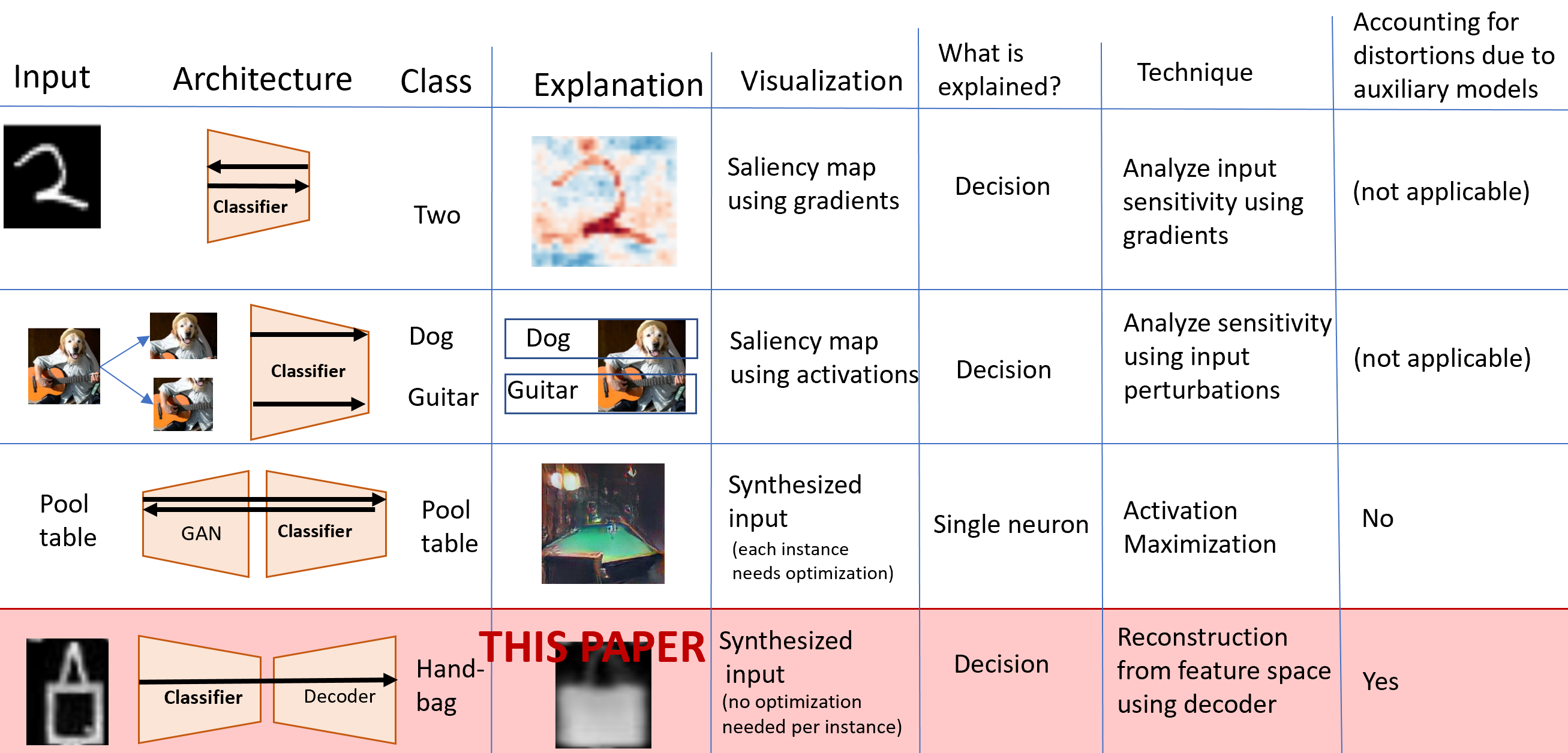}
  \caption{Method Overview. Figures are from cited papers.} \label{fig:met}
\end{figure*}

We discuss approaches that allow to visualize single features as well as to understand particular decisions summarized in Figure \ref{fig:met}.% We also show examples of saliency maps to contrast to our work.

\section{Note on theory}
From a practical point of view, there is limited value in explaining a linear regression model with few variables, since linear regression models are transparent. However, for more complex (linear) models involving many, potentially transformed input attributes $x_i$ such as $\sin(x_i)$, $x_i^2$, explanations might still be helpful. Furthermore, linear regression exhibits nice properties for theoretical analysis and it is widely used.